%% file: main.tex
\documentclass{article} 
\usepackage{iclr2025_conference,times}

\input{math_commands.tex}

\usepackage{comment}

\usepackage{amsmath}
\usepackage{tikz}
\usepackage{pgfplots}\usepgfplotslibrary{groupplots,units}\pgfplotsset{compat=1.18}
\usetikzlibrary{decorations}

\usepackage{hyperref}
\usepackage{url}
\usepackage[most]{tcolorbox}

\title{An Extensive Evaluation of PDDL Capabilities in off-the-shelf LLMs}

\iclrfinalcopy

\author{Kaustubh Vyas\textsuperscript{\,$\dagger$}, Damien Graux\textsuperscript{\,$\dagger$}, S{\'e}bastien Montella\textsuperscript{\,$\dagger$}, Pavlos Vougiouklis\textsuperscript{\,$\dagger$}, \\
\textbf{Ruofei Lai\textsuperscript{\,$\dagger$}, Keshuang Li\textsuperscript{\,$\dagger$}, Yang Ren\textsuperscript{\,$\dagger$}, \& Jeff Z. Pan\textsuperscript{\,$\dagger$\,,\,\P}}\\
\textsuperscript{$\dagger$}~Huawei Research Ltd., \textsuperscript{\P}~University of Edinburgh, United Kingdom\\
\texttt{damien.graux@huawei.com},~\texttt{j.z.pan@ed.ac.uk} \\
}

\begin{document}

\maketitle

\begin{abstract}
In recent advancements, large language models (LLMs) have exhibited proficiency in code generation and chain-of-thought reasoning, laying the groundwork for tackling automatic formal planning tasks.
This study evaluates the potential of LLMs to understand and generate Planning Domain Definition Language (PDDL), an essential representation in artificial intelligence planning.
We conduct an extensive analysis across 20 distinct models spanning 7 major LLM families, both commercial and open-source. 
Our comprehensive evaluation sheds light on the zero-shot LLM capabilities of parsing, generating, and reasoning with PDDL.
Our findings indicate that while some models demonstrate notable effectiveness in handling PDDL, others pose limitations in more complex scenarios requiring nuanced planning knowledge.
These results highlight the promise and current limitations of LLMs in formal planning tasks, offering insights into their application and guiding future efforts in AI-driven planning paradigms.
\end{abstract}

\section{Introduction}

Automated planning has long been a cornerstone of artificial intelligence, traditionally relying on explicit domain knowledge encoded in formal languages such as PDDL. In recent years, the rapid evolution of large language models (LLMs) has sparked considerable interest in their ability to bridge the gap between natural language descriptions and formal planning representations. 

Early studies by \cite{zuo2024planetarium} and \cite{oswald2024large} demonstrated that LLMs are capable of translating natural language descriptions into syntactically valid PDDL representations. However, these pioneering works also revealed significant gaps, as the generated planning domains frequently diverge from gold-standard models, both syntactically and semantically. This observation has spurred further research into the underlying reasoning capabilities of LLMs and their potential role in executing complete planning tasks

Advancement in LLMs fuelled recent efforts that looked into how these multi-billion parameter models can be best employed as agents~\cite{huang2024understanding}. Building on this momentum, several strategies have been proposed to map user instructions into PDDL problems \cite{ijcai2023p839,Liu2023LLMPEL,dagan2023dynamicplanningllm,gestrin2024towards,zhang-etal-2024-pddlego}, without however providing conclusive evidence for the feasibility of the task in the general domain. These studies underscore both the promise and the challenges inherent in leveraging LLMs for complex planning and reasoning tasks, where transforming natural language into an executable agentic workflow remains a non-trivial endeavor.

In this study, we step back to examine the fluency of twenty LLMs from seven major families in the PDDL language, focusing on their ability to parse, generate, and reason with PDDL. Specifically, we leverage the Planetarium benchmark~\cite{zuo2024planetarium} alongside the dataset introduced by \citet{oswald2024large} to assess how well these models understand and generate actions, problems, and plans. By analyzing a randomly sampled subset of over 13,000 (NL-instruction, PDDL-problem) pairs, our results show that although some models demonstrate moderate proficiency in handling PDDL, the majority struggle to convert natural language instructions into fully correct PDDL representations. This challenge is especially evident in smaller LLMs, which often fail to produce parsable PDDL.

It is important to note that our evaluation focuses exclusively on zero-shot prompting, as our aim is to test the planning capabilities of off-the-shelf LLMs. Although few-shot prompting approaches might further improve performance, they were not considered in this study to maintain a clear assessment of the models in their unmodified state.

Our contributions are twofold: (1) an extensive evaluation of the zero-shot planning performance of LLMs as reflected in their PDDL handling, and (2) an investigation into the feasibility of employing these models as co-pilots in planning tasks.

\section{Extensive PDDL capability evaluation}
From a high-level point of view, PDDL involves three types of elements: the \textbf{domains} to represent the possible \textit{actions} available in a certain space, the \textbf{problems} which roughly encode the premise and the goal of a \textit{real world} operation to be performed in a defined space (\textit{i.e.} domain) and finally the \textbf{plans} that represent the effective set of \textit{actions} to be run to perform the \textit{real world} operation, achieving the \textit{goal}.

Therefore, practically, we stressed the considered LLMs to generate all or part of the aforementioned elements, while maintaining a wide set of evaluation scores across the involved steps to fuel the discussion and draw conclusions.

\subsection{Action Generation}
\textbf{Task Signature} = \texttt{[input:} NL instruction, PDDL domain predicates; 
\texttt{output:} PDDL action\texttt{]} \\ 
\\
%
%
We rely on the benchmark proposed by~\cite{oswald2024large} to evaluate the action generation capabilities: given a seed domain file and the NL description of an action, we let the LLM generate it in proper PDDL syntax. In their article, the authors shared a set of 32 NL-to-Action instructions distributed across 9 popular PDDL domains. We enriched these by generating 4 NL-variations for each NL-to-instruction pair to obtain a dataset of $160$ $\left[(1+4)\times 32\right]$ instructions. \\
To assess the results, we score along the following dimensions: \textbf{Parsable}: Determines if the output conforms to correct PDDL syntax. \textbf{Solvable}: Measures how well the action integrates into the target domain (\textit{e.g.} the action may be syntactically correct but involving type mismatches, wrong number of variables for some predicates,\ldots). \textbf{Equivalent}: Syntactically valid PDDL that integrates with the desired domain under the domain equivalence heuristic.\\
To measure the similarity between the generated action and the gold standard, we calculate the normalised differences in their preconditions and effects, and then subtract this value from one to derive a similarity score.
\[
\text{Similarity} = 1 - \frac{|\text{A}_{\text{pre}} \Delta \widehat{\text{A}}_{\text{pre}}| + |\text{A}_{\text{ef}} \Delta \widehat{\text{A}}_{\text{ef}}|}{|\text{A}_{\text{pre}} \cup \widehat{\text{A}}_{\text{pre}}| + |\text{A}_{\text{ef}} \cup \widehat{\text{A}}_{\text{ef}}|}
\]
Where \(\text{A}_{\text{pre}}\), \(\text{A}_{\text{ef}}\) are preconditions and effects in the gold action and \(\widehat{\text{A}}_{\text{pre}}\), \(\widehat{\text{A}}_{\text{ef}}\) are preconditions and effects in the LLM generated action.

\subsection{Problem Generation} 
\textbf{Task Signature} = \texttt{[input:} NL instruction, PDDL domain; \texttt{output:} PDDL problem\texttt{]} \\
\\
We choose the Planetarium benchmark to evaluate the problem generation capabilities of models~\cite{zuo2024planetarium}. The benchmark was primarily selected due its size that enabled a comprehensive evaluation on our side. In particular, we randomly selected 10\% of the full dataset, resulting in a test set consisting of 13\,203 (NL-instruction, PDDL-problem) pairs. Metrics for this set of experiments are as above (with a slight difference): \textbf{Parsable}: generated PDDL adheres to the syntactic rules of the language, \textbf{Solvable}: the generated problem can be effectively processed by existing PDDL planners, reflecting its practical utility, \textbf{Equivalent}: matches the gold standard in both structure and semantics. \\
When it comes to measure the similarity between the gold and the generated PDDL problem, we use ChrF \cite{popovic-2015-chrf} as it is a standard metric to evaluate code generation tasks \cite{evtikhiev2023out}. By employing the ChrF metric, we can objectively assess subtle differences between the generated and reference PDDL code, offering a nuanced understanding of each model’s translation accuracy.


\subsection{Plan Generation}  
\textbf{Task Signature} = \texttt{[input:} PDDL domain, PDDL problem; \texttt{output:} Plan\texttt{]}\\
\\
Finally, although LLMs are not expected to outperform conventional planners\footnote{See~\citet{frances2017purely} for an extensive evaluation of PDDL planners, which also showcases the planner used as reference in our study: BFWS-FF~\cite{lipovetzky2017best,lipovetzky2017polynomial}.}—since their reasoning capabilities rely on intrinsic parametric knowledge rather than explicit logical reasoning\footnote{\cite{mirzadeh2024gsm} provides a more nuanced understanding of LLMs' capabilities and limitations in (mathematical) reasoning.}—we also aimed to assess their ability to \textit{plan in PDDL} when provided with pairs of domain and problem. For this purpose, we selected domain-problem pairs from the Planetarium benchmark to prompt the models for plan generation. To evaluate generalisation, we categorised these problems based on their level of abstractness, classifying descriptions as either \textbf{explicit} or \textbf{abstract}. Explicit descriptions are direct propositions found in the PDDL problem (e.g., “block1 is on block2”), whereas abstract descriptions summarise a state (e.g., “all blocks are in a single tower”). Because these descriptions encapsulate both the initial and goal states, there are four possible categories: (1) Abstract initial and goal states, (2) Abstract initial but explicit goal, (3) Explicit initial but abstract goal and (4) Explicit initial and goal states. In total, we selected 40 representatives from each category, yielding 160 pairs. The correctness of the generated plans is then verified using VAL\footnote{\url{https://github.com/KCL-Planning/VAL}}, a tool that assesses whether a plan is compatible with the specified PDDL domain and problem.

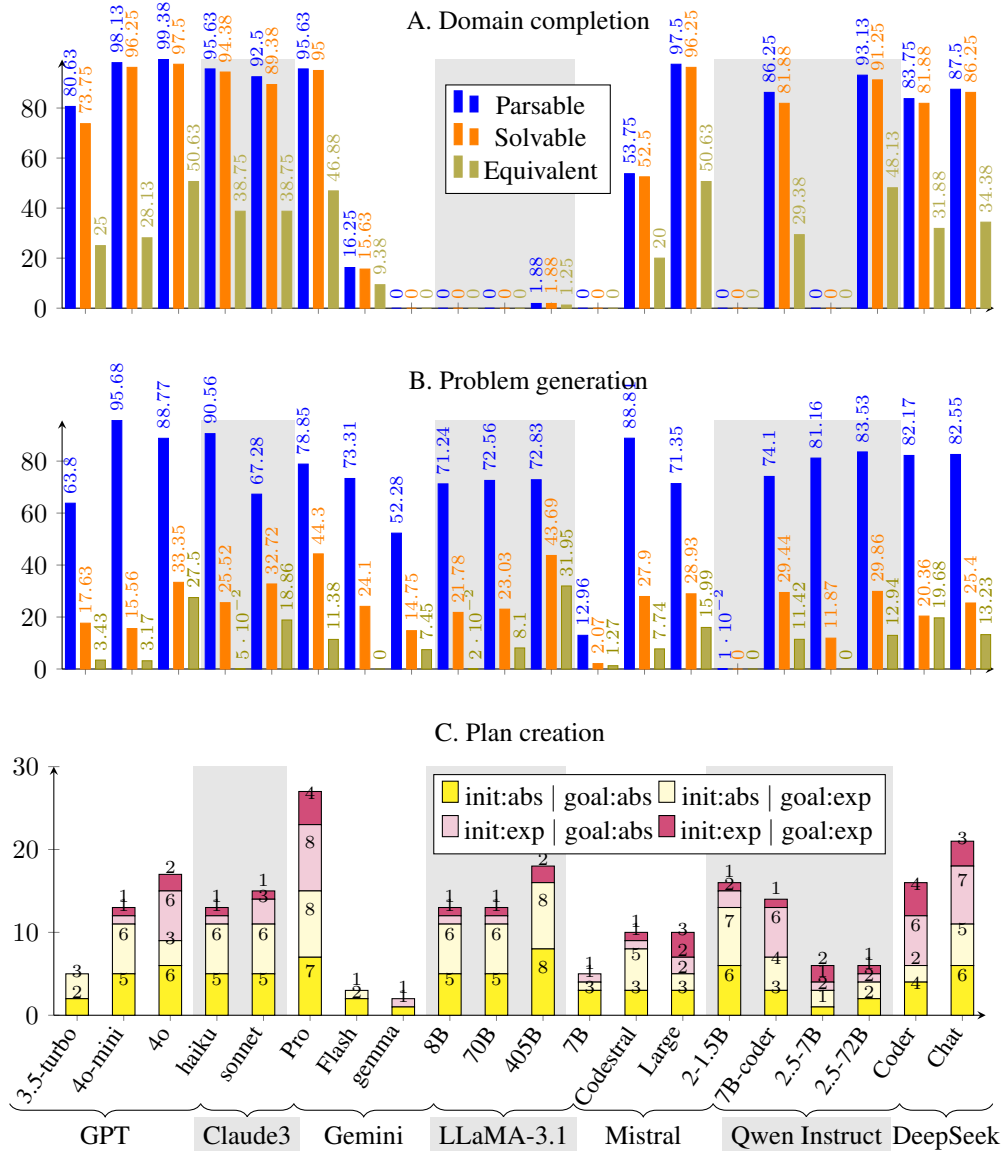
\begin{figure}[t]
\centering
\begin{tikzpicture}
\begin{axis}[
    title={A. Domain completion},
    width=1\textwidth,
    height=0.35\textwidth,
    bar width=0.13cm,
    ybar,
    enlargelimits=0.2,
    legend style={at={(0.5,0.9)},anchor=north,legend columns=1},
    xtick={1,2,3,4,5,6,7,8,9,10,11,12,13,14,15,16,17,18,19,20},
    xticklabels=\empty,
    xticklabel style={rotate=45, anchor=east, font=\footnotesize, name=T\ticknum},
    xmin = 0.5, xmax = 20.5, ymin=0,
    axis lines=left,
    nodes near coords,
    every node near coord/.append style={rotate=90,font=\scriptsize},
    nodes near coords align={horizontal},
    axis background/.style={%
        preaction={
            path picture={
                \fill[fill=black!10,line width=0mm] (axis cs:3.5,0) rectangle (axis cs:5.5,100.5);
                \fill[fill=black!10,line width=0mm] (axis cs:8.5,0) rectangle (axis cs:11.5,100.5);
                \fill[fill=black!10,line width=0mm] (axis cs:14.5,0) rectangle (axis cs:18.5,100.5);
    }}}
]
\addplot[blue,fill=blue] coordinates {(1 , 80.63) (2 , 98.13) (3 , 99.38) (4 , 95.63) (5 , 92.50) (6 , 95.63) (7 , 16.25) (8 , 0.00) (9 , 0.00) (10 , 0.00) (11 , 1.88) (12 , 0.00) (13 , 53.75) (14 ,97.50) (15 , 0.00) (16 , 86.25) (17 , 0.00) (18 , 93.13) (19 , 83.75) (20 , 87.50)}; 
\addplot[orange,fill=orange] coordinates {(1 , 73.75) (2 , 96.25) (3 , 97.50) (4 , 94.38) (5 , 89.38) (6 , 95.00) (7 , 15.63) (8 , 0.00) (9 , 0.00) (10 , 0.00) (11 , 1.88) (12 , 0.00) (13 , 52.50) (14 ,96.25) (15 , 0.00) (16 , 81.88) (17 , 0.00) (18 , 91.25) (19 , 81.88) (20 , 86.25)}; 
\addplot[olive!70,fill=olive!70] coordinates {(1 , 25.00) (2 , 28.13) (3 , 50.63) (4 , 38.75) (5 , 38.75) (6 , 46.88) (7 , 9.38) (8 , 0.00) (9 , 0.00) (10 , 0.00) (11 , 1.25) (12 , 0.00) (13 , 20.00) (14 , 50.63) (15 , 0.00) (16 , 29.38) (17 , 0.00) (18 , 48.13) (19 , 31.88) (20 , 34.38)}; 
\legend{Parsable,Solvable,Equivalent}
\end{axis}
\end{tikzpicture}
\begin{tikzpicture}
\begin{axis}[
    title={B. Problem generation},
    width=1\textwidth,
    height=0.35\textwidth,
    bar width=0.13cm,
    ybar,
    enlargelimits=0.2,
    legend style={at={(0.5,1)},anchor=north,legend columns=2},
    xtick={1,2,3,4,5,6,7,8,9,10,11,12,13,14,15,16,17,18,19,20},
    xticklabels=\empty,
    xticklabel style={rotate=45, anchor=east, font=\footnotesize, name=T\ticknum},
    xmin = 0.5, xmax = 20.5, ymin=0,
    axis lines=left,
    nodes near coords,
    every node near coord/.append style={rotate=90,font=\scriptsize},
    nodes near coords align={horizontal},
    axis background/.style={%
        preaction={
            path picture={
                \fill[fill=black!10,line width=0mm] (axis cs:3.5,0) rectangle (axis cs:5.5,100.5);
                \fill[fill=black!10,line width=0mm] (axis cs:8.5,0) rectangle (axis cs:11.5,100.5);
                \fill[fill=black!10,line width=0mm] (axis cs:14.5,0) rectangle (axis cs:18.5,100.5);
    }}}
]
\addplot[blue,fill=blue] coordinates {(1 , 63.80) (2 , 95.68) (3 , 88.77) (4 , 90.56) (5 , 67.28) (6 , 78.85) (7 , 73.31) (8 , 52.28) (9 , 71.24) (10 , 72.56) (11 , 72.83) (12 , 12.96) (13 , 88.81) (14 , 71.35) (15 , 0.01) (16 , 74.10) (17 , 81.16) (18 , 83.53) (19 , 82.17) (20 , 82.55)}; 
\addplot[orange,fill=orange] coordinates {(1 , 17.63) (2 , 15.56) (3 , 33.35) (4 , 25.52) (5 , 32.72) (6 , 44.30) (7 , 24.10) (8 , 14.75) (9 , 21.78) (10 , 23.03) (11 , 43.69) (12 , 2.07) (13 , 27.90) (14 , 28.93) (15 , 0.00) (16 , 29.44) (17 , 11.87) (18 , 29.86) (19 , 20.36) (20 , 25.40)}; 
\addplot[olive,fill=olive!70] coordinates {(1 , 3.43) (2 , 3.17) (3 , 27.50) (4 , 0.05) (5 , 18.86) (6 , 11.38) (7 , 0.00) (8 , 7.45) (9 , 0.02) (10 , 8.10) (11 , 31.95) (12 , 1.27) (13 , 7.74) (14 , 15.99) (15 , 0.00) (16 , 11.42) (17 , 0.00) (18 , 12.94) (19 , 19.68) (20 , 13.23)}; 
\end{axis}
\end{tikzpicture}
\begin{tikzpicture}
\begin{axis}[
    title={C. Plan creation},
    width=1\textwidth,
    height=0.35\textwidth,
    bar width=0.3cm,
    ybar stacked,
    enlargelimits=0.15,
    legend style={at={(0.65,0.97)},anchor=north,legend columns=2},
    xtick={1,2,3,4,5,6,7,8,9,10,11,12,13,14,15,16,17,18,19,20},
    xticklabels={3.5-turbo,4o-mini,4o,haiku,sonnet,Pro,Flash,gemma,8B,70B,405B,7B,Codestral,Large,2-1.5B,7B-coder,2.5-7B,2.5-72B,Coder,Chat},
    xticklabel style={rotate=55, anchor=east, font=\footnotesize, name=T\ticknum},
    xmin = 0.5, xmax = 20.5, ymin=0, ymax=30,
    axis lines=left,
    nodes near coords,
    every node near coord/.append style={font=\scriptsize},
    nodes near coords align={vertical},
    axis background/.style={%
        preaction={
            path picture={
                \fill[fill=black!10,line width=0mm] (axis cs:3.5,0) rectangle (axis cs:5.5,100.5);
                \fill[fill=black!10,line width=0mm] (axis cs:8.5,0) rectangle (axis cs:11.5,100.5);
                \fill[fill=black!10,line width=0mm] (axis cs:14.5,0) rectangle (axis cs:18.5,100.5);
    }}}
]
\addplot[fill=yellow!90] coordinates {(1 , 2) (2 , 5) (3 , 6) (4 , 5) (5 , 5) (6 , 7) (7 , 2) (8 , 1) (9 , 5) (10 , 5) (11 , 8) (12 , 3) (13 , 3) (14 , 3) (15 , 6) (16 , 3) (17 , 1) (18 , 2) (19 , 4) (20 , 6)}; 
\addplot[fill=yellow!20] coordinates {(1 , 3) (2 , 6) (3 , 3) (4 , 6) (5 , 6) (6 , 8) (7 , 1) (8 , 0) (9 , 6) (10 , 6) (11 , 8) (12 , 1) (13 , 5) (14 , 2) (15 , 7) (16 , 4) (17 , 2) (18 , 2) (19 , 2) (20 , 5)}; 
\addplot[fill=purple!20] coordinates {(1 , 0) (2 , 1) (3 , 6) (4 , 1) (5 , 3) (6 , 8) (7 , 0) (8 , 1) (9 , 1) (10 , 1) (11 , 0) (12 , 1) (13 , 1) (14 , 2) (15 , 2) (16 , 6) (17 , 1) (18 , 1) (19 , 6) (20 , 7)}; 
\addplot[fill=purple!70] coordinates {(1 , 0) (2 , 1) (3 , 2) (4 , 1) (5 , 1) (6 , 4) (7 , 0) (8 , 0) (9 , 1) (10 , 1) (11 , 2) (12 , 0) (13 , 1) (14 , 3) (15 , 1) (16 , 1) (17 , 2) (18 , 1) (19 , 4) (20 , 3)}; 
\legend{init:abs $|$ goal:abs,init:abs $|$ goal:exp,init:exp $|$ goal:abs,init:exp $|$ goal:exp}
\end{axis}
\begin{scope}[decoration=brace]
  \pgfdecorationsegmentamplitude=5pt
  \draw[decorate] ([yshift=(-1.1cm),xshift=(0.4cm)]T2.east) -- ([yshift=(-1.1cm),xshift=(-0.9cm)]T0.east) node[midway,below=\pgfdecorationsegmentamplitude] {GPT};
  \draw[decorate] ([yshift=(-1.1cm),xshift=(0.4cm)]T4.east) -- ([yshift=(-1.1cm),xshift=(0.4cm)]T2.east) node[midway,fill=black!10,below=\pgfdecorationsegmentamplitude] {Claude3};
  \draw[decorate] ([yshift=(-1.1cm),xshift=(0.4cm)]T7.east) -- ([yshift=(-1.1cm),xshift=(0.4cm)]T4.east) node[midway,below=\pgfdecorationsegmentamplitude] {Gemini};
  \draw[decorate] ([yshift=(-1.1cm),xshift=(0.4cm)]T10.east) -- ([yshift=(-1.1cm),xshift=(0.4cm)]T7.east) node[midway,fill=black!10,below=\pgfdecorationsegmentamplitude] {LLaMA-3.1};
  \draw[decorate] ([yshift=(-1.1cm),xshift=(0.4cm)]T13.east) -- ([yshift=(-1.1cm),xshift=(0.4cm)]T10.east) node[midway,below=\pgfdecorationsegmentamplitude] {Mistral};
  \draw[decorate] ([yshift=(-1.1cm),xshift=(0.4cm)]T17.east) -- ([yshift=(-1.1cm),xshift=(0.4cm)]T13.east) node[midway,fill=black!10,below=\pgfdecorationsegmentamplitude] {Qwen Instruct};
  \draw[decorate] ([yshift=(-1.1cm),xshift=(0.4cm)]T19.east) -- ([yshift=(-1.1cm),xshift=(0.4cm)]T17.east) node[midway,below=\pgfdecorationsegmentamplitude] {DeepSeek};
\end{scope}
\end{tikzpicture}
\caption{LLM performances across the three benchmarks (\textit{higher the better}).}
\label{fig:performance}
\end{figure}

\subsection{Considered LLMs}

To review the capabilities of language models to deal with PDDL, we utilised LLMs from several leading organisations, ensuring that both general-purpose and specialist models (i.e. chatting, code generation or instruction-following modes) are considered.
%
Our set of models includes LLMs from
\begin{itemize}
\item \textbf{OpenAI} (\textit{GPT-3.5-turbo}, \textit{GPT-4o-mini}, \textit{GPT-4o}), 
\item \textbf{Anthropic} (\textit{Claude-3-Haiku} and \textit{Claude-3-Sonnet}),
\item \textbf{Google} (\textit{Gemini-1.5-Pro}, \textit{Gemini-1.5-Flash} and \textit{Gemma-2-9B-it}),
\item \textbf{Meta} (\textit{LLaMA-3.1-8B-Instruct}, \textit{LLaMA-3.1-70B-Instruct}, and \textit{LLaMA3.1-405B-Instruct}),
\item \textbf{Mistral} (\textit{Large2}, \textit{7B-Instruct}, and \textit{Codestral}),
\item \textbf{DeepSeek} (\textit{Coder-V2} and \textit{Chat-V2}),
\item \textbf{Alibaba} (\textit{Qwen2-1.5B-Instruct}, \textit{Qwen2.5-7B-Instruct}, \textit{Qwen2.5-Coder-7B-Instruct}, and \textit{Qwen2.5-72B-Instruct}).
\end{itemize}
%
Overall, this set involves members of \textbf{7} distinct providers, including commercial and open LLMs. In addition, this set allows us to compare behaviors and performances across different parameter numbers and specialities.

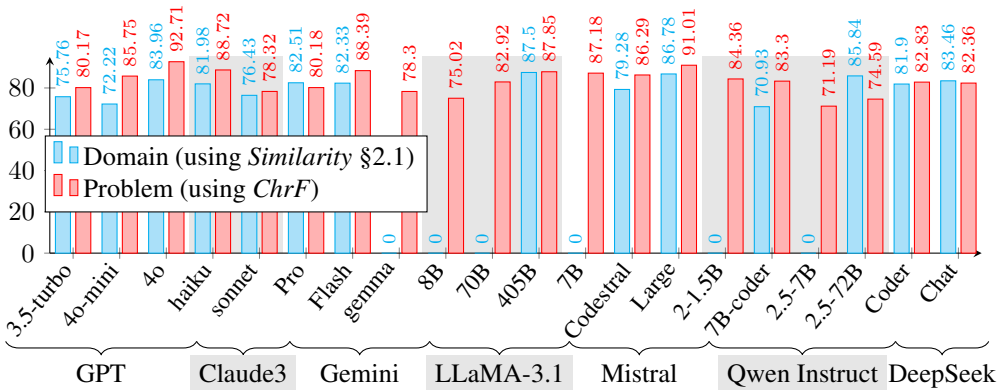
\begin{figure}
\begin{tikzpicture}
\begin{axis}[
    width=1\textwidth,
    height=0.3\textwidth,
    bar width=0.2cm,
    ybar,
    enlargelimits=0.15,
    legend style={at={(0.2,0.6)},anchor=north,legend columns=1},
    legend cell align={left},
    xtick={1,2,3,4,5,6,7,8,9,10,11,12,13,14,15,16,17,18,19,20},
    xticklabels={3.5-turbo,4o-mini,4o,haiku,sonnet,Pro,Flash,gemma,8B,70B,405B,7B,Codestral,Large,2-1.5B,7B-coder,2.5-7B,2.5-72B,Coder,Chat},
    xticklabel style={rotate=50, anchor=east, font=\footnotesize, name=T\ticknum},
    xmin = 0.5, xmax = 20.5, ymin=0, ymax=95,
    axis lines=left,
    nodes near coords,
    every node near coord/.append style={rotate=90,font=\scriptsize},
    nodes near coords align={horizontal},
    axis background/.style={%
        preaction={
            path picture={
                \fill[fill=black!10,line width=0mm] (axis cs:3.5,0) rectangle (axis cs:5.5,100.5);
                \fill[fill=black!10,line width=0mm] (axis cs:8.5,0) rectangle (axis cs:11.5,100.5);
                \fill[fill=black!10,line width=0mm] (axis cs:14.5,0) rectangle (axis cs:18.5,100.5);
    }}}
]
\addplot[cyan,fill=cyan!30] coordinates {(1 , 75.76) (2 , 72.22) (3 , 83.96) (4 , 81.98) (5 , 76.43) (6 , 82.51) (7 , 82.33) (8 , 0.00) (9 , 0.00) (10 , 0.0) (11 , 87.5) (12 , 0.0) (13 , 79.28) (14 , 86.78) (15 , 0.00) (16 , 70.93) (17 , 0.00) (18 , 85.84) (19 , 81.90) (20 , 83.46)}; 
\addplot coordinates {(1 , 80.17) (2 , 85.75) (3 , 92.71) (4 , 88.72) (5 , 78.32) (6 , 80.18) (7 , 88.39) (8 , 78.30) (9 , 75.02) (10 , 82.92) (11 , 87.85) (12 , 87.18) (13 , 86.29) (14 , 91.01) (15 , 84.36) (16 , 83.30) (17 , 71.19) (18 , 74.59) (19 , 82.83) (20 , 82.36)};  
\legend{Domain (using \textit{Similarity} \S2.1),Problem (using \textit{ChrF})}
\end{axis}
\begin{scope}[decoration=brace]
  \pgfdecorationsegmentamplitude=5pt
  \draw[decorate] ([yshift=(-1.1cm),xshift=(0.4cm)]T2.east) -- ([yshift=(-1.1cm),xshift=(-0.9cm)]T0.east) node[midway,below=\pgfdecorationsegmentamplitude] {GPT};
  \draw[decorate] ([yshift=(-1.1cm),xshift=(0.4cm)]T4.east) -- ([yshift=(-1.1cm),xshift=(0.4cm)]T2.east) node[midway,fill=black!10,below=\pgfdecorationsegmentamplitude] {Claude3};
  \draw[decorate] ([yshift=(-1.1cm),xshift=(0.4cm)]T7.east) -- ([yshift=(-1.1cm),xshift=(0.4cm)]T4.east) node[midway,below=\pgfdecorationsegmentamplitude] {Gemini};
  \draw[decorate] ([yshift=(-1.1cm),xshift=(0.4cm)]T10.east) -- ([yshift=(-1.1cm),xshift=(0.4cm)]T7.east) node[midway,fill=black!10,below=\pgfdecorationsegmentamplitude] {LLaMA-3.1};
  \draw[decorate] ([yshift=(-1.1cm),xshift=(0.4cm)]T13.east) -- ([yshift=(-1.1cm),xshift=(0.4cm)]T10.east) node[midway,below=\pgfdecorationsegmentamplitude] {Mistral};
  \draw[decorate] ([yshift=(-1.1cm),xshift=(0.4cm)]T17.east) -- ([yshift=(-1.1cm),xshift=(0.4cm)]T13.east) node[midway,fill=black!10,below=\pgfdecorationsegmentamplitude] {Qwen Instruct};
  \draw[decorate] ([yshift=(-1.1cm),xshift=(0.4cm)]T19.east) -- ([yshift=(-1.1cm),xshift=(0.4cm)]T17.east) node[midway,below=\pgfdecorationsegmentamplitude] {DeepSeek};
\end{scope}
\end{tikzpicture}
\caption{Performances of LLMs as co-pilots, reviewing closeness (\%) of generations to the ``gold''.}
\label{fig:assisting-performance}
\end{figure}

\subsection{Results}

\paragraph*{Domain (Fig.\ref{fig:performance}A)} We review the performance of 20 distinct LLMs in populating PDDL domain files with new actions based on NL instructions. 
While most LLMs perform well in generating correct actions, a notable decline in performance is observed in their ability to produce equivalent actions. Among the models, \textit{GPT-4o}, \textit{Qwen2.5-72B-instruct}, and \textit{Mistral-Large2} stand out as the top performers across all three metrics for action generation. In contrast, some models consistently fail to adhere to the required action syntax. Notably, the entire \textit{LLaMA} family performs very poorly, irrespective of the number of parameters.
It is worthwhile noting that all the models have their respective \textit{parsable} and \textit{solvable} values very close, in other words the difference between them for a model is of few points. This implies that not only models are able to generate proper PDDL syntax (\textit{parsable}) but they comply with the given domain (\textit{solvable}), such a property could be useful if the models were used as assistants, see Figure~\ref{fig:assisting-performance} and its associated discussion for more details.


\paragraph*{Problem (Fig.\ref{fig:performance}B)} We next assessed the ability of LLMs to generate complete PDDL problems based on NL instructions and a corresponding PDDL domain. While most models can produce syntactically correct and parsable PDDL problems, their performance declines significantly when required to solve the problem for generating a plan or producing problems equivalent to the gold. Interestingly, the \textit{LLaMA} family, which struggled with syntax in action evaluations, demonstrates improved accuracy for this task. Even though \textit{LLaMA-3.1-405B} achieves the highest equivalency rate at about 32\%, its performance --and that of all off-the-shelf models-- remains inadequate for this task, suggesting that further techniques such as few-shot prompting, fine-tuning, or other enhancements are necessary to assist PDDL problem generation, as explored by~\cite{zuo2024planetarium} for instance.
Furthermore, models are not consistent through the three metrics: for instance \textit{GPT-40-mini} which has the bext score for \textit{parsable} ends up being in the worst when it comes to \textit{equivalent}, a similar behaviour goes for \textit{Codestral} too.

\paragraph*{Plan (Fig.\ref{fig:performance}C)} Investigating the plan generation, as expected, we find that given a PDDL domain and problem, the models struggle to generate PDDL plans. This was tested on 160 data points, with \textit{Gemini-1.5-pro} performing the best, yet achieving a valid plan in only 16.87\% of the cases. In contrast, the BFWS-FF planner succeeded in generating a conclusive plan 86.25\% of the time. Additionally, we explored the plan generation relative to the abstractness of the initial state and goal of the PDDL problem. Across all LLMs, we observe that they perform better when the initial state of the PDDL problem is abstract, though, no such pattern is noticed regarding the abstractness of the goal.
Once again, the models are not performing similarly across the different tasks. Typically, our smallest model in the mix (\textit{Qwen2-1.5B-Instruct}) which had scores almost all null for the domain and problem tasks, happens to be in the top-5 for the plan generation.

\subsection{Discussions}

Our results indicate that while most LLMs can generate syntactically correct PDDL, they largely lack the capacity to generate effective problems and plans for addressing the input instructions. This behaviour becomes increasingly evident as the complexity of the experimental setup rises.



\paragraph*{Parameter Number} With the exception of the \textit{LLaMa} family, an increased number of parameters in LLMs does not consistently lead to better PDDL fluency. This suggests that current pre- and post-training approaches are not effectively scaling these models to meet the demands of complex PDDL generation, including, but not limited to, long-horizon planning tasks. A clear challenge emerges with models like the \textit{LLaMA} family (see Fig.\ref{fig:performance}), which, while effective in problem generation, struggle significantly with action generation. This issue stems from a syntactic bias: for example, instead of the correct keyword ``\texttt{:precondition}'', they generate ``\texttt{:preconditions}'', making actions unparsable. Similar patterns are observed in \textit{Gemma-2-9B-it}, \textit{Mistral-7B-Instruct}, or \textit{Qwen2.5-72B-Instruct}, which incorrectly output ``\texttt{(action\ldots)}'' rather than ``\texttt{(:action\ldots)}''. Similarly, while generating problems, models often struggle with maintaining correct action sequences in the goal, occasionally placing semantically similar actions in the goal that are not defined in the domain, leading to mismatches. 

\paragraph*{LLMs as copilots} However, our findings, in Fig.\ref{fig:assisting-performance}, reveal that despite these challenges, LLM-generated PDDL actions and problems demonstrate a high degree of closeness to the gold standard. This suggests that, while the models (off-the-shelf) may not yet be fully reliable for independent use, they hold strong potential as supportive tools. By generating near-accurate PDDL structures, these models can serve as co-pilots, streamlining the drafting process and allowing experts to focus on refinement and optimization rather than building from the ground up. The position articulated in \citet{pmlr-v235-kambhampati24a} underscores that, although LLMs may not inherently plan effectively, they can nonetheless play a significant supportive role in LLM-modulo planning frameworks. This further reinforces our findings, highlighting the viability of these models as valuable assistants in structured problem-solving tasks.


\begin{table}[t]
\centering
\caption{Costs (US\$) of the 3 experiments, \textsc{na} when model-API is free. Horizon scores: (\%) for the problem equivalence.}
\label{tab:cost-horizon}
\begin{tabular}{|c||c||ccc|}\hline
\textbf{LLM} & \textbf{Cost} & \textbf{Short} & \textbf{Medium} & \textbf{Long} \\\hline
\textit{GPT-3.5-turbo} &13.28&20.98&3.09&1.13\\
\textit{GPT-4o-mini} &5.16&1.13&3.23&3.19\\
\textit{GPT-4o} &83.16&20.42&28.08&11.32\\\hline
\textit{Claude-3-Haiku} &12.15&0.36&0.04&0.0\\
\textit{Claude-3-Sonnet} &118.25&15.52&19.29&6.42\\\hline
\textit{Gemini-1.5-Pro} &46.19&5.71&11.98&1.51\\
\textit{Gemini-1.5-Flash} &3.28&0.0&0.0&0.0\\
\textit{Gemma-2-9B-it} &1.22&4.24&7.71&1.13\\\hline
\textit{LLaMA-3.1-8B-Instruct} &0.89&0&0.02&0\\
\textit{LLaMA-3.1-70B-Instruct} &21.81&0&8.58&1.51\\
\textit{LLaMA-3.1-405B-Instruct} &86.13&32.27&33.29&10.65\\\hline
\textit{Mistral-7B-Instruct} &\textsc{na}&3.15&1.23&1.13\\
\textit{Codestral} &6.68&5.43&6.72&4.74\\
\textit{Mistral-Large2} &71.51&11.27&16.03&35.5\\\hline
\textit{Qwen2-1.5B-Instruct} &0.32&0.0&0.0&0.0\\
\textit{Qwen2.5-Coder-7B-Instruct} &\textsc{na}&27.44&13.15&0.38\\
\textit{Qwen2.5-7B-Instruct} &\textsc{na}&0.0&0.0&0.0\\
\textit{Qwen2.5-72B-Instruct} &8.84&13.48&13.49&4.45\\\hline
\textit{DeepSeek-Coder-V2} &3.24&11.83&20.79&2.26\\
\textit{DeepSeek-Chat-V2} &3.24&6.55&14.05&1.88\\\hline
\end{tabular}
\end{table}

\paragraph*{Cost} Moreover, as highlighted in Table~\ref{tab:cost-horizon}, these performances are not correlated with the various LLM costs. Indeed, overall, looking at the aggregated costs for the whole experiments, we cannot say that most expensive models lead to best results, see for instance the cases of Claude-3-Sonnet (\$118) and Qwen2.5-72B-Instruct (\$8.8) which have similar scale of performances. However, we can see that \textbf{within} an LLM-family, the more expensive models usually allow for better results, see \textit{e.g.} OpenAI's models where GPT-4o is strictly better than the other two.

\paragraph*{Horizon Distance} In order to explore how the number of actions (context) in a PDDL problem might affect the generation, in Table \ref{tab:cost-horizon}, we analyse the performance of models with three horizons: \textbf{Short} (up to 10 actions), \textbf{Medium} (from 11 to 50) and \textbf{Long} (more than 50). While short horizons perform decently as they are straightforward to process, they often lack the necessary context to provide sufficient information for optimal model performance. In contrast, medium horizons offer a richer and more comprehensive context, enabling the model to achieve results that are not only comparable to those with short horizons but often surpass them (Table \ref{tab:cost-horizon}). This suggests that medium horizons strike a balance by providing enough information to enhance generation for more complex queries without overwhelming the models. 

The extended context and complexity of PDDL problems exacerbate these issues, as models find it challenging to sustain coherence and alignment across the entire generation process, as scores decrease with long-horizon planning. As described in~\cite{chen2024backbone}, long-horizon planning tasks involve ``extended sequences of actions'' or span ``over a prolonged period of time''. Reasoning over longer context windows has been a traditional challenge associated with language models, especially in tasks requiring understanding of complex, long-range dependencies \cite{li2024longcontext}. We find that this behaviour exacerbates in the context of generating longer, expressive problems. We believe that problem decomposition strategies, that would seek to split and hierarchically construct an expected problem in a bottom-up fashion could offer promising directions for future work.


\section{Related Work --- PDDL $\cap$ LLM}

The generation of PDDL domains and problems has recently garnered significant attention as a means to enhance planning via large language models (LLMs) \cite{Strobel_2020,silver2020pddlgym,silver2022pddl,vyas2025hive}. In parallel, the advent of sophisticated prompting techniques has unlocked new applications for LLMs \cite{survey-prompting-2023,graux2024prompteng}. Nonetheless, while LLMs have demonstrated planning capabilities \cite{huang2024understanding}, they continue to struggle with long-horizon planning, uncertainty in generated plans, and generalisation to unseen domains \cite{robovqa2023arxiv}. Consequently, several works have aimed to bridge the gap between the probabilistic nature of LLMs and the deterministic requirements of PDDL-based planners. For instance, \citet{collinsWong2022} compared the out-of-distribution robustness of PDDL-augmented LLMs with human reasoning, highlighting clear limitations in current LLM approaches.

In many settings, LLMs have proven more effective at translating natural language into formal representations rather than performing the planning itself, as noted in works such as \citet{htn2pddl,Helmert2009ConciseFR,xie2023translatingnaturallanguageplanning}. This observation has spurred strategies that decompose the problem into translating user instructions into PDDL problems, solving these problems via formal logic within the PDDL framework, and then translating the resulting plans back into natural language \cite{ijcai2023p839,Liu2023LLMPEL,dagan2023dynamicplanningllm,silver2024generalized,gestrin2024towards,mahdavi2024leveragingenvironmentinteractionautomated,zhang-etal-2024-pddlego}.

More recent contributions have further refined the dialogue between LLMs and planning. \cite{hao-etal-2023-reasoning} propose that reasoning with a language model can be reinterpreted as planning with an integrated world model, while \cite{rossetti2024learning} explore the learning of general policies for planning directly via GPT models. In addition, benchmark efforts such as PlanBench introduced by Valmeekam et al. \cite{valmeekam2023planbench} and critical investigations into LLM planning abilities \cite{valmeekam2023planbench} provide valuable insights into the performance and limitations of current models.

Novel benchmarks such as PlanBench \cite{valmeekam2023planbench}, AutoPlanBench \cite{AutoPlanBench}, Planetarium \cite{zuo2024planetarium}, and the domain benchmark from \citet{oswald2024large} have been introduced to assess LLMs’ planning capabilities using PDDL. However, to the best of our knowledge, the recent families of foundational models have not yet been extensively benchmarked to reveal their inherent robustness and reliability in handling PDDL generation. In this study, we explore the capacity of these foundational models to generate both PDDL domains and problems, thereby extending prior evaluations and situating our work alongside the latest advances in planning with LLMs.

\section{Conclusion}

In this study, we experimentally reviewed the PDDL capabilities of a large panel of language models: twenty in total, representing multiple dimensions of the current state-of-the-art.
Our evaluations show that (some) LLMs can be used to generate actions to complete PDDL domains, they may also be used to assist in the task of generating PDDL problems from NL instructions. However, as expected, they reveal being poor planners and it is better to rely on ``real'' planners which have been developed by the community for decades.
Surprisingly also, behaviours given a specific model are not uniform across tasks as good performers (even leaders) for a certain PDDL aspect may turn out to be among the worst ones later.
Overall, we hope to pave the road to future efforts in AI-driven planning challenges.

\bibliography{bibliography}
\bibliographystyle{iclr2025_conference}

\appendix

\section{Examples of 0-shot prompts}

In order to give a better idea on the prompts we triggered at the language models, we provide in this Appendix an example for each of the tested task, \textit{i.e.} Action creation, Problem generation and Plan creation. We refer the reader to the main body of the article for more details, see \textit{e.g.} Figure~\ref{fig:performance} (A, B and C) for details of the results\footnote{Moreover, we will provide --once not \textit{double-blind} anymore-- a link toward our Github repository which gathers all the benchmarks (including the prompts) that were used to assess the LLM performances.}
. In particular, here we provide:
\begin{itemize}
    \item The action generation (\texttt{put-down}) for the \texttt{BLOCKS} domain;
    \item A problem generation in \texttt{blocksworld} domain (\texttt{init}=\texttt{holding\_one});
    \item A plan generation in \texttt{blocksworld} domain (\texttt{init}=\texttt{on\_table}, \texttt{goal}=\texttt{tower}).
\end{itemize}

\begin{tcolorbox}[
    breakable,
    enhanced,
    colback=white!100,
    colframe=black!50!blue,
    width=\columnwidth,
    subtitle style={boxrule=0.4pt,colback=yellow!50!red!25!white},
    title= \textbf{Prompt and result for the action generation (\texttt{put-down}) for the \texttt{BLOCKS} domain}]
    \phantomsection
Given a description of an action in some domain, convert it to Planning Domain Definition Language (PDDL) action. You may only use the allowed predicates for each action.\\Allowed Predicates:\\- (handempty) : the hand is empty.\quad~- (holding ?x - block): block x is held.\\- (clear ?x - block): block x is clear.\quad- (on ?x - block ?y - block) : block x is on block y.\\- (ontable ?x - block): block x is on the table.\\Input: The action, ``put-down'' will have the hand put down a block.\\PDDL Action:
\tcblower
(~:action put-down\\\strut\qquad:parameters (?x - block)\\\strut\qquad:precondition (holding ?x)\\\strut\qquad:effect (and (not (holding ?x)) (clear ?x) (handempty) (ontable ?x))\qquad)
\end{tcolorbox}

\begin{tcolorbox}[
    breakable,
    enhanced,
    colback=white!100,
    colframe=black!50!blue,
    width=\columnwidth,
    subtitle style={boxrule=0.4pt,colback=yellow!50!red!25!white},
    title= \textbf{A problem generation in \texttt{blocksworld} domain (\texttt{init}=\texttt{holding\_one})}]
    \phantomsection
Provide me with the complete, valid problem PDDL file that describes the following planning problem directly without further explanations or texts. The domain for the planning problem is:\\(define (domain blocksworld) (:requirements :strips) (:predicates (arm-empty) (clear ?x) (holding ?x) (on ?x ?y) (on-table ?x)) (:action pickup   :parameters (?ob)   :precondition (and (clear ?ob) (on-table ?ob) (arm-empty))   :effect (and (holding ?ob) (not (clear ?ob)) (not (on-table ?ob)) (not (arm-empty))) ) (:action putdown   :parameters (?ob)   :precondition (holding ?ob)   :effect (and (clear ?ob) (arm-empty) (on-table ?ob) (not (holding ?ob))) ) (:action stack   :parameters (?ob ?underob)   :precondition (and (clear ?underob) (holding ?ob))   :effect (and (arm-empty) (clear ?ob) (on ?ob ?underob) (not (clear ?underob)) (not (holding ?ob))) ) (:action unstack   :parameters (?ob ?underob)   :precondition (and (on ?ob ?underob) (clear ?ob) (arm-empty))   :effect (and (holding ?ob) (clear ?underob) (not (on ?ob ?underob)) (not (clear ?ob)) (not (arm-empty))) ))\\Planning problem:\\You have 17 blocks. You are holding b1. b2 is clear. b2 is on the table. b3 is clear. b3 is on the table. b4 is clear. b4 is on the table. b5 is clear. b5 is on the table. b6 is clear. b6 is on the table. b7 is clear. b7 is on the table. b8 is clear. b8 is on the table. b9 is clear. b9 is on the table. b10 is clear. b10 is on the table. b11 is clear. b11 is on the table. b12 is clear. b12 is on the table. b13 is clear. b13 is on the table. b14 is clear. b14 is on the table. b15 is clear. b15 is on the table. b16 is clear. b16 is on the table. b17 is clear. b17 is on the table.\\Your goal is to stack the blocks into a towers of heights 1, 1, 1, 2, 3, 9.\\Problem PDDL:
\tcblower
(define (problem holding\_one\_to\_tower\_1\_1\_1\_2\_3\_9)\\\strut\quad (:domain blocksworld)\\\strut\quad (:requirements :strips)\\\strut\quad (:objects b1 b10 b11 b12 b13 b14 b15 b16 b17 b2 b3 b4 b5 b6 b7 b8 b9)\\\strut\quad (:init (clear b10) (clear b11) (clear b12) (clear b13) (clear b14) (clear b15) (clear b16) (clear b17) (clear b2) (clear b3) (clear b4) (clear b5) (clear b6) (clear b7) (clear b8) (clear b9) (holding b1) (on-table b10) (on-table b11) (on-table b12) (on-table b13) (on-table b14) (on-table b15) (on-table b16) (on-table b17) (on-table b2) (on-table b3) (on-table b4) (on-table b5) (on-table b6) (on-table b7) (on-table b8) (on-table b9))\\\strut\quad (:goal (and (arm-empty) (clear b1) (on-table b1) (clear b2) (on-table b2) (clear b3) (on-table b3) (clear b4) (on b4 b5) (on-table b5) (clear b6) (on b6 b7) (on b7 b8) (on-table b8) (clear b9) (on b9 b10) (on b10 b11) (on b11 b12) (on b12 b13) (on b13 b14) (on b14 b15) (on b15 b16) (on b16 b17) (on-table b17)))\qquad)
\end{tcolorbox}

\begin{tcolorbox}[
    breakable,
    enhanced,
    colback=white!100,
    colframe=black!50!blue,
    width=\columnwidth,
    subtitle style={boxrule=0.4pt,colback=yellow!50!red!25!white},
    title= \textbf{Plan generation in \texttt{blocksworld} domain (\texttt{init}=\texttt{on\_table}, \texttt{goal}=\texttt{tower})}]
    \phantomsection
Given a PDDL domain and a PDDL problem file, come up with the plan associated with the problem. The domain describes the possible actions and their effects, while the problem file details the specific scenario to be solved. Do not generate anything but the correct plan\\Domain PDDL:\\(define (domain blocksworld) (:requirements :strips) (:predicates (arm-empty) (clear ?x)  (holding ?x)  (on ?x ?y)  (on-table ?x)) (:action pickup     :parameters (?ob)     :precondition (and (clear ?ob) (on-table ?ob) (arm-empty))     :effect (and (holding ?ob) (not (clear ?ob)) (not (on-table ?ob)) (not (arm-empty))) )  (:action putdown     :parameters (?ob)     :precondition (holding ?ob)     :effect (and (clear ?ob) (arm-empty) (on-table ?ob) (not (holding ?ob))) )  (:action stack     :parameters (?ob ?underob)     :precondition (and (clear ?underob) (holding ?ob))     :effect (and (arm-empty) (clear ?ob) (on ?ob ?underob) (not (clear ?underob)) (not (holding ?ob))) )  (:action unstack     :parameters (?ob ?underob)     :precondition (and (on ?ob ?underob) (clear ?ob) (arm-empty))     :effect (and (holding ?ob) (clear ?underob) (not (on ?ob ?underob)) (not (clear ?ob)) (not (arm-empty))) ))\\Problem PDDL:\\(define (problem on\_table\_to\_tower\_1\_1\_1\_1\_3\_13)(:domain blocksworld)(:requirements :strips)(:objects b1 b10 b11 b12 b13 b14 b15 b16 b17 b18 b19 b2 b20 b3 b4 b5 b6 b7 b8 b9)(:init (arm-empty) (clear b1) (clear b10) (clear b11) (clear b12) (clear b13) (clear b14) (clear b15) (clear b16) (clear b17) (clear b18) (clear b19) (clear b2) (clear b20) (clear b3) (clear b4) (clear b5) (clear b6) (clear b7) (clear b8) (clear b9) (on-table b1) (on-table b10) (on-table b11) (on-table b12) (on-table b13) (on-table b14) (on-table b15) (on-table b16) (on-table b17) (on-table b18) (on-table b19) (on-table b2) (on-table b20) (on-table b3) (on-table b4) (on-table b5) (on-table b6) (on-table b7) (on-table b8) (on-table b9))(:goal (and (arm-empty) (clear b1) (on-table b1) (clear b2) (on-table b2) (clear b3) (on-table b3) (clear b4) (on-table b4) (clear b5) (on b5 b6) (on b6 b7) (on-table b7) (clear b8) (on b8 b9) (on b9 b10) (on b10 b11) (on b11 b12) (on b12 b13) (on b13 b14) (on b14 b15) (on b15 b16) (on b16 b17) (on b17 b18) (on b18 b19) (on b19 b20) (on-table b20))))\\Plan:
\tcblower
(define (problem on\_table\_to\_tower\_1\_1\_1\_1\_3\_13)\\\strut\quad (:domain blocksworld)\qquad (:requirements :strips)\\\strut\quad (:objects b1 b10 b11 b12 b13 b14 b15 b16 b17 b18 b19 b2 b20 b3 b4 b5 b6 b7 b8 b9)\\\strut\quad (:init (arm-empty) (clear b1) (clear b10) (clear b11) (clear b12) (clear b13) (clear b14) (clear b15) (clear b16) (clear b17) (clear b18) (clear b19) (clear b2) (clear b20) (clear b3) (clear b4) (clear b5) (clear b6) (clear b7) (clear b8) (clear b9) (on-table b1) (on-table b10) (on-table b11) (on-table b12) (on-table b13) (on-table b14) (on-table b15) (on-table b16) (on-table b17) (on-table b18) (on-table b19) (on-table b2) (on-table b20) (on-table b3) (on-table b4) (on-table b5) (on-table b6) (on-table b7) (on-table b8) (on-table b9))\\\strut\quad (:goal (and (arm-empty) (clear b1) (on-table b1) (clear b2) (on-table b2) (clear b3) (on-table b3) (clear b4) (on-table b4) (clear b5) (on b5 b6) (on b6 b7) (on-table b7) (clear b8) (on b8 b9) (on b9 b10) (on b10 b11) (on b11 b12) (on b12 b13) (on b13 b14) (on b14 b15) (on b15 b16) (on b16 b17) (on b17 b18) (on b18 b19) (on b19 b20) (on-table b20)))\qquad)
\end{tcolorbox}

\end{document}

%% file: math_commands.tex
\usepackage{amsmath,amsfonts,bm}

\def\eqref#1{equation~\ref{#1}}

\def\1{\bm{1}}

\DeclareMathAlphabet{\mathsfit}{\encodingdefault}{\sfdefault}{m}{sl}
\SetMathAlphabet{\mathsfit}{bold}{\encodingdefault}{\sfdefault}{bx}{n}













%% file: main.bbl
\begin{thebibliography}{34}
\providecommand{\natexlab}[1]{#1}
\providecommand{\url}[1]{\texttt{#1}}
\expandafter\ifx\csname urlstyle\endcsname\relax
  \providecommand{\doi}[1]{doi: #1}\else
  \providecommand{\doi}{doi: \begingroup \urlstyle{rm}\Url}\fi

\bibitem[Alford et~al.(2009)Alford, Kuter, and Nau]{htn2pddl}
Ronald Alford, Ugur Kuter, and Dana Nau.
\newblock Translating htns to pddl: A small amount of domain knowledge can go a long way.
\newblock pp.\  1629--1634, 01 2009.

\bibitem[Chen et~al.(2024)Chen, Chen, Lee, Ge, Rojas, and Kormushev]{chen2024backbone}
Xiaoshuai Chen, Wei Chen, Dongmyoung Lee, Yukun Ge, Nicolas Rojas, and Petar Kormushev.
\newblock A backbone for long-horizon robot task understanding.
\newblock \emph{arXiv:2408.01334}, 2024.

\bibitem[Collins et~al.(2022)Collins, Wong, Feng, Wei, and Tenenbaum]{collinsWong2022}
Katherine~M. Collins, Catherine Wong, Jiahai Feng, Megan Wei, and Joshua~B. Tenenbaum.
\newblock Structured, flexible, and robust: benchmarking and improving large language models towards more human-like behavior in out-of-distribution reasoning tasks, 2022.
\newblock URL \url{https://arxiv.org/abs/2205.05718}.

\bibitem[Dagan et~al.(2023)Dagan, Keller, and Lascarides]{dagan2023dynamicplanningllm}
Gautier Dagan, Frank Keller, and Alex Lascarides.
\newblock Dynamic planning with a llm, 2023.
\newblock URL \url{https://arxiv.org/abs/2308.06391}.

\bibitem[Evtikhiev et~al.(2023)Evtikhiev, Bogomolov, Sokolov, and Bryksin]{evtikhiev2023out}
Mikhail Evtikhiev, Egor Bogomolov, Yaroslav Sokolov, and Timofey Bryksin.
\newblock Out of the bleu: how should we assess quality of the code generation models?
\newblock \emph{Journal of Systems and Software}, 203:\penalty0 111741, 2023.

\bibitem[Frances et~al.(2017)Frances, Ram{\'\i}rez~J{\'a}vega, Lipovetzky, and Geffner]{frances2017purely}
Guillem Frances, Miquel Ram{\'\i}rez~J{\'a}vega, Nir Lipovetzky, and Hector Geffner.
\newblock Purely declarative action descriptions are overrated: Classical planning with simulators.
\newblock In \emph{26th Int. Joint Conf. on Artificial Intelligence; Aug 19-25; Melbourne, Australia. p. 4294-301.} IJCAI Organization, 2017.

\bibitem[Gestrin et~al.(2024)Gestrin, Kuhlmann, and Seipp]{gestrin2024towards}
Elliot Gestrin, Marco Kuhlmann, and Jendrik Seipp.
\newblock Towards robust {LLM}-driven planning from minimal text descriptions.
\newblock In \emph{Workshop on Human-Aware Explainable Planning}, 2024.
\newblock URL \url{https://openreview.net/forum?id=NmzHuV101q}.

\bibitem[Graux et~al.(2024)Graux, Montella, Jabeen, Gardent, and Pan]{graux2024prompteng}
Damien Graux, S{\'e}bastien Montella, Hajira Jabeen, Claire Gardent, and Jeff~Z Pan.
\newblock {[PromptEng] First International Workshop on Prompt Engineering for Pre-Trained Language Models}.
\newblock In \emph{Companion Proceedings of the ACM on Web Conference 2024}, pp.\  1311--1312, 2024.

\bibitem[Hao et~al.(2023)Hao, Gu, Ma, Hong, Wang, Wang, and Hu]{hao-etal-2023-reasoning}
Shibo Hao, Yi~Gu, Haodi Ma, Joshua Hong, Zhen Wang, Daisy Wang, and Zhiting Hu.
\newblock Reasoning with language model is planning with world model.
\newblock In \emph{EMNLP}, pp.\  8154--8173, Singapore, December 2023.
\newblock \doi{10.18653/v1/2023.emnlp-main.507}.

\bibitem[Helmert(2009)]{Helmert2009ConciseFR}
Malte Helmert.
\newblock Concise finite-domain representations for pddl planning tasks.
\newblock \emph{Artif. Intell.}, 173:\penalty0 503--535, 2009.
\newblock URL \url{https://api.semanticscholar.org/CorpusID:9377590}.

\bibitem[Huang et~al.(2024)Huang, Liu, Chen, Wang, Wang, Lian, Wang, Tang, and Chen]{huang2024understanding}
Xu~Huang, Weiwen Liu, Xiaolong Chen, Xingmei Wang, Hao Wang, Defu Lian, Yasheng Wang, Ruiming Tang, and Enhong Chen.
\newblock Understanding the planning of {LLM} agents: A survey.
\newblock \emph{arXiv preprint arXiv:2402.02716}, 2024.

\bibitem[Kambhampati et~al.(2024)Kambhampati, Valmeekam, Guan, \ldots, and B~Murthy]{pmlr-v235-kambhampati24a}
Subbarao Kambhampati, Karthik Valmeekam, Lin Guan, \ldots, and Anil B~Murthy.
\newblock Position: {LLM}s can’t plan, but can help planning in {LLM}-modulo frameworks.
\newblock In \emph{ICML}, volume 235, pp.\  22895--22907. PMLR, 21--27 Jul 2024.

\bibitem[Li et~al.(2024)Li, Zhang, Do, Yue, and Chen]{li2024longcontext}
Tianle Li, Ge~Zhang, Quy~Duc Do, Xiang Yue, and Wenhu Chen.
\newblock Long-context llms struggle with long in-context learning, 2024.

\bibitem[Lipovetzky \& Geffner(2017{\natexlab{a}})Lipovetzky and Geffner]{lipovetzky2017best}
Nir Lipovetzky and Hector Geffner.
\newblock Best-first width search: Exploration and exploitation in classical planning.
\newblock In \emph{Proceedings of the AAAI Conference on Artificial Intelligence}, volume~31, 2017{\natexlab{a}}.

\bibitem[Lipovetzky \& Geffner(2017{\natexlab{b}})Lipovetzky and Geffner]{lipovetzky2017polynomial}
Nir Lipovetzky and Hector Geffner.
\newblock A polynomial planning algorithm that beats lama and ff.
\newblock In \emph{Proceedings of the International Conference on Automated Planning and Scheduling}, volume~27, pp.\  195--199, 2017{\natexlab{b}}.

\bibitem[Liu et~al.(2023{\natexlab{a}})Liu, Jiang, Zhang, Liu, Zhang, Biswas, and Stone]{Liu2023LLMPEL}
B.~Liu, Yuqian Jiang, Xiaohan Zhang, Qian Liu, Shiqi Zhang, Joydeep Biswas, and Peter Stone.
\newblock Llm+p: Empowering large language models with optimal planning proficiency.
\newblock \emph{ArXiv}, abs/2304.11477, 2023{\natexlab{a}}.
\newblock URL \url{https://api.semanticscholar.org/CorpusID:258298051}.

\bibitem[Liu et~al.(2023{\natexlab{b}})Liu, Yuan, Fu, Jiang, Hayashi, and Neubig]{survey-prompting-2023}
Pengfei Liu, Weizhe Yuan, Jinlan Fu, Zhengbao Jiang, Hiroaki Hayashi, and Graham Neubig.
\newblock Pre-train, prompt, and predict: A systematic survey of prompting methods in natural language processing.
\newblock \emph{ACM Comput. Surv.}, 55\penalty0 (9), jan 2023{\natexlab{b}}.
\newblock ISSN 0360-0300.
\newblock \doi{10.1145/3560815}.
\newblock URL \url{https://doi.org/10.1145/3560815}.

\bibitem[Mahdavi et~al.(2024)Mahdavi, Aoki, Tang, and Cao]{mahdavi2024leveragingenvironmentinteractionautomated}
Sadegh Mahdavi, Raquel Aoki, Keyi Tang, and Yanshuai Cao.
\newblock Leveraging environment interaction for automated {PDDL} generation and planning with large language models, 2024.
\newblock URL \url{https://arxiv.org/abs/2407.12979}.

\bibitem[Mirzadeh et~al.(2024)Mirzadeh, Alizadeh, Shahrokhi, Tuzel, Bengio, and Farajtabar]{mirzadeh2024gsm}
Iman Mirzadeh, Keivan Alizadeh, Hooman Shahrokhi, Oncel Tuzel, Samy Bengio, and Mehrdad Farajtabar.
\newblock {GSM}-symbolic: Understanding the limitations of mathematical reasoning in large language models.
\newblock \emph{arXiv preprint arXiv:2410.05229}, 2024.

\bibitem[Oswald et~al.(2024)Oswald, Srinivas, Kokel, Lee, Katz, and Sohrabi]{oswald2024large}
James Oswald, Kavitha Srinivas, Harsha Kokel, Junkyu Lee, Michael Katz, and Shirin Sohrabi.
\newblock {Large Language Models as Planning Domain Generators}.
\newblock In \emph{Proceedings of the Int. Conference on Automated Planning and Scheduling}, volume~34, pp.\  423--431, 2024.

\bibitem[Pallagani et~al.(2023)Pallagani, Muppasani, Srivastava, Rossi, Horesh, Murugesan, Loreggia, Fabiano, Joseph, and Kethepalli]{ijcai2023p839}
Vishal Pallagani, Bharath Muppasani, Biplav Srivastava, Francesca Rossi, Lior Horesh, Keerthiram Murugesan, Andrea Loreggia, Francesco Fabiano, Rony Joseph, and Yathin Kethepalli.
\newblock Plansformer tool: Demonstrating generation of symbolic plans using transformers.
\newblock In Edith Elkind (ed.), \emph{{IJCAI-23}}, pp.\  7158--7162. IJCAI Organization, 8 2023.
\newblock \doi{10.24963/ijcai.2023/839}.

\bibitem[Popovi{\'c}(2015)]{popovic-2015-chrf}
Maja Popovi{\'c}.
\newblock chr{F}: character n-gram {F}-score for automatic {MT} evaluation.
\newblock In \emph{Tenth Workshop on Statistical Machine Translation}, pp.\  392--395, Lisbon, Portugal, September 2015. Association for Computational Linguistics.
\newblock \doi{10.18653/v1/W15-3049}.

\bibitem[Rossetti et~al.(2024)Rossetti, Tummolo, Gerevini, Putelli, Serina, Chiari, and Olivato]{rossetti2024learning}
Nicholas Rossetti, Massimiliano Tummolo, Alfonso~Emilio Gerevini, Luca Putelli, Ivan Serina, Mattia Chiari, and Matteo Olivato.
\newblock Learning general policies for planning through gpt models.
\newblock In \emph{ICAPS}, volume~34, pp.\  500--508, 2024.

\bibitem[Sermanet et~al.(2023)Sermanet, Ding, Zhao, Xia, \ldots, and Cao]{robovqa2023arxiv}
Pierre Sermanet, Tianli Ding, Jeffrey Zhao, Fei Xia, \ldots, and Yuan Cao.
\newblock Robovqa: Multimodal long-horizon reasoning for robotics.
\newblock In \emph{arXiv preprint arXiv:2311.00899}, 2023.

\bibitem[Silver \& Chitnis(2020)Silver and Chitnis]{silver2020pddlgym}
Tom Silver and Rohan Chitnis.
\newblock Pddlgym: Gym environments from pddl problems.
\newblock In \emph{Int. Conference on Automated Planning and Scheduling (ICAPS) PRL Workshop}, 2020.
\newblock URL \url{https://github.com/tomsilver/pddlgym}.

\bibitem[Silver et~al.(2022)Silver, Hariprasad, Shuttleworth, Kumar, Lozano-P{\'e}rez, and Kaelbling]{silver2022pddl}
Tom Silver, Varun Hariprasad, Reece~S Shuttleworth, Nishanth Kumar, Tom{\'a}s Lozano-P{\'e}rez, and Leslie~Pack Kaelbling.
\newblock {PDDL} planning with pretrained large language models.
\newblock In \emph{NeurIPS 2022 Foundation Models for Decision Making Workshop}, 2022.
\newblock URL \url{https://openreview.net/forum?id=1QMMUB4zfl}.

\bibitem[Silver et~al.(2024)Silver, Dan, Srinivas, Tenenbaum, Kaelbling, and Katz]{silver2024generalized}
Tom Silver, Soham Dan, Kavitha Srinivas, Josh Tenenbaum, Leslie Kaelbling, and Michael Katz.
\newblock Generalized planning in {PDDL} domains with pretrained large language models.
\newblock In \emph{AAAI Conference on Artificial Intelligence (AAAI)}, 2024.

\bibitem[Stein et~al.(2024)Stein, Fišer, Hoffmann, and Koller]{AutoPlanBench}
Katharina Stein, Daniel Fišer, Jörg Hoffmann, and Alexander Koller.
\newblock Autoplanbench: Automatically generating benchmarks for llm planners from pddl, 2024.
\newblock URL \url{https://arxiv.org/abs/2311.09830}.

\bibitem[Strobel \& Kirsch(2020)Strobel and Kirsch]{Strobel_2020}
Volker Strobel and Alexandra Kirsch.
\newblock \emph{MyPDDL: Tools for Efficiently Creating PDDL Domains and Problems}, pp.\  67–90.
\newblock Springer International Publishing, 2020.
\newblock ISBN 9783030385613.
\newblock URL \url{http://dx.doi.org/10.1007/978-3-030-38561-3_4}.

\bibitem[Valmeekam et~al.(2023)Valmeekam, Marquez, Olmo, Sreedharan, and Kambhampati]{valmeekam2023planbench}
Karthik Valmeekam, Matthew Marquez, Alberto Olmo, Sarath Sreedharan, and Subbarao Kambhampati.
\newblock Planbench: An extensible benchmark for evaluating large language models on planning and reasoning about change, 2023.

\bibitem[Vyas et~al.(2025)Vyas, Graux, Yang, Montella, Diao, Zhou, Vougiouklis, Lai, Ren, Li, and Pan]{vyas2025hive}
Kaustubh Vyas, Damien Graux, Yijun Yang, Sébastien Montella, Chenxin Diao, Wendi Zhou, Pavlos Vougiouklis, Ruofei Lai, Yang Ren, Keshuang Li, and Jeff~Z. Pan.
\newblock {From An LLM Swarm To A PDDL-Empowered HIVE: Planning Self-Executed Instructions In A Multi-Modal Jungle}.
\newblock \emph{International Conference on Learning Representations (ICLR)}, 2025.

\bibitem[Xie et~al.(2023)Xie, Yu, Zhu, Bai, Gong, and Soh]{xie2023translatingnaturallanguageplanning}
Yaqi Xie, Chen Yu, Tongyao Zhu, Jinbin Bai, Ze~Gong, and Harold Soh.
\newblock Translating natural language to planning goals with large-language models, 2023.
\newblock URL \url{https://arxiv.org/abs/2302.05128}.

\bibitem[Zhang et~al.(2024)Zhang, Jansen, Zhang, Clark, Callison-Burch, and Tandon]{zhang-etal-2024-pddlego}
Li~Zhang, Peter Jansen, Tianyi Zhang, Peter Clark, Chris Callison-Burch, and Niket Tandon.
\newblock {PDDLEGO}: Iterative planning in textual environments.
\newblock In Danushka Bollegala and Vered Shwartz (eds.), \emph{13th Joint Conf. on Lexical and Computational Semantics (*SEM 2024)}, pp.\  212--221, Mexico City, Mexico, June 2024. Association for Computational Linguistics.
\newblock \doi{10.18653/v1/2024.starsem-1.17}.
\newblock URL \url{https://aclanthology.org/2024.starsem-1.17}.

\bibitem[Zuo et~al.(2024)Zuo, Velez, Li, Littman, and Bach]{zuo2024planetarium}
Max Zuo, Francisco~Piedrahita Velez, Xiaochen Li, Michael~L Littman, and Stephen~H Bach.
\newblock {Planetarium: A Rigorous Benchmark for Translating Text to Structured Planning Languages}.
\newblock \emph{arXiv:2407.03321}, 2024.

\end{thebibliography}
